\documentclass{article}
\usepackage{spconf,amsmath,graphicx}
\usepackage{subfigure}
\newtheorem{theorem}{Theorem}

\title{A Discrete Scheme for Computing Image's Weighted Gaussian Curvature}
\name{Yuanhao Gong\thanks{This work is supported by National Natural Science Foundation (61907031).}, Wenming Tang, Lebin Zhou, Lantao Yu, Guoping Qiu}
\address{College of Electronics and Information Engineering, Shenzhen University, China, gong@szu.edu.cn}
\begin{document}
%
\maketitle
\begin{abstract}
Weighted Gaussian Curvature is an important measurement for images. However, its conventional computation scheme has low performance, low accuracy and requires that the input image must be second order differentiable. To tackle these three issues, we propose a novel discrete computation scheme for the weighted Gaussian curvature. Our scheme does not require the second order differentiability. Moreover, our scheme is more accurate, has smaller support region and computationally more efficient than the conventional schemes. Therefore, our scheme holds promise for a large range of applications where the weighted Gaussian curvature is needed, for example, image smoothing, cartoon texture decomposition, optical flow estimation, etc.
\end{abstract}
\section{Introduction}
\label{sec:intro}
Smoothness of images is an important measurement for image processing tasks. It can be used for evaluating image processing algorithms' performance. For example, in image smoothing, the details of input images are removed by processing algorithms. Evaluating the smoothed results by some smoothness measurement is an important topic. In image denoising, smoothness measurement can be adopted as an evaluation metric for the resulting clear images. In image decomposition, it can also be used to indicate the smoothness of the base layer (cartoon part).   

There are several types of smoothness measurements in the literature. A classical example is the well known Total Variation (TV), which seeks to minimize the gradient field of the input image \cite{TV1992}. Such first order regularization leads to piece wise constant outputs. Thanks to its effectiveness and convexity property, TV has been used in a large range of applications, including denoising, smoothing, image reconstruction and optical flow estimation. 

Another typical example of smoothness measurements is the curvature, which requires second order derivatives. The classical mean curvature  and Gaussian curvature can be adopted as smoothness measurements \cite{gong:phd,meanZhu}. 

Mean curvature has been used as regularization for ill-posed image processing tasks \cite{meanZhu,mean1997,gong:Bernstein,Zhu2007}. Minimizing mean curvature is assuming that the ground truth has zero mean curvature everywhere on the surface. In differential geometry, it has been proved that such surfaces must be minimal surfaces. Minimal surfaces have minimal area for given boundary condition. Meanwhile, such minimal area is also related with the bending energy in physics. Recent achievements on mean curvature can be found in \cite{gong:cf,2020Image,GONG2019329}.

Gaussian curvature is another important second order smoothness measurement for surfaces in differential geometry. The surfaces with zero Gaussian curvature everywhere are called developable, which means that they can be mapped onto a plane without any distortion (the distance and angle are preserved during the mapping). Such property can benefit a wide range of applications in image processing.

\subsection{Gaussian Curvature}
Let $\vec{x}=(x,y)^T$ denote the spatial coordinate and $I(\vec{x}):R^2\rightarrow R$ be the image. We can embed the image into the three dimensional space, forming a surface in 3D. More specifically, the image surface of $I(\vec{x})$ is $\vec{\Psi}=(\vec{x},I(\vec{x}))$. The normal vector of $\vec{\Psi}$ is given by $\vec{n}=\frac{(-I_x,-I_y,1)}{\sqrt{1+I_x^2+I_y^2}}$, where $I_x=\frac{\partial I}{\partial x}$ and $I_y=\frac{\partial I}{\partial y}$. The first fundamental form of $\vec{\Psi}$ is defined as 
\begin{equation}
	F=\begin{pmatrix} 1+I_x^2 & I_{xy} \\ I_{yx }& 1+I_y^2 \end{pmatrix}\,.
\end{equation} The second fundamental form of $\vec{\Psi}$ is defined as
\begin{equation}
D=\begin{pmatrix} \frac{\partial \vec{\Psi}}{\partial xx}\cdot\vec{n} & \frac{\partial \vec{\Psi}}{\partial xy}\cdot\vec{n} \\ \frac{\partial \vec{\Psi}}{\partial yx}\cdot\vec{n} & \frac{\partial \vec{\Psi}}{\partial yy}\cdot\vec{n} \end{pmatrix}\,.
\end{equation}
Then, the Gaussian curvature on the surface $\vec{\Psi}$ is defined as
\begin{equation}
\label{eq:gc}
	K(I(\vec{x}))=\frac{\mathrm{det}(D)}{\mathrm{det}(F)}=\frac{I_{xx}I_{yy}-I_{xy}^2}{(1+I_x^2+I_y^2)^2}\,.
\end{equation}

\subsection{Limitations of Gaussian Curvature}
Although Gaussian curvature is well investigated in differential geometry, it has limitations in image processing tasks \cite{firsov:2006,mesh2006,Lu2011}. First, the digital discrete image intensity values in neighbor pixels might have a jump (sharp edges) and thus it might not be second order differentiable. Such discontinuity leads to the issue for computing Gaussian curvature by Eq.~\ref{eq:gc}.

Second, the Gaussian curvature is not robust to image contrast change. Let  another image $\hat{I}(\vec{x})$ be just a simple scaling of $I(\vec{x})$, that is $\hat{I}(\vec{x})=sI(\vec{x})$, where $s>0$ is a scalar. Then the Gaussian curvature of $\hat{I}$ is
 \begin{equation}
 K(\hat{I}(\vec{x}))=\frac{s^2(I_{xx}I_{yy}-I_{xy}^2)}{(1+s^2(I_x^2+I_y^2))^2}\,.
 \end{equation} When the gradient magnitude is close to zero (flat region, $I_x^2+I_y^2\approx 0$), $K(\hat{I})\approx s^2K(I)$. When the gradient magnitude is much larger than one (sharp edges, $I_x^2+I_y^2>> 1$), $K(\hat{I})\approx \frac{1}{s^2}K(I)$. Therefore, the Gaussian curvature is not robust with image contrast change for both flat regions and edges.
 
Third, the Gaussian curvature is not robust to pixel size (spatial resolution). In the discrete case, the Gaussian curvature can be computed by finite difference schemes
\begin{align}
I_{x}\approx &(I(x+h,y)-I(x,y))/h\,, \\
I_{y}\approx &	(I(x,y+h)-I(x,y))/h\,, \\
I_{xx}\approx &(I(x+h,y)+I(x-h,y)-I(x,y))/h^2\,, \\
I_{yy}\approx &	(I(x,y+h)+I(x,y-h)-I(x,y))/h^2\,, \\
I_{xy}\approx &	(I(x+h,y+h)+I(x-h,y-h))/h^2 \nonumber \\
&-(I(x+h,y-h)+I(x-h,y+h))/h^2\,, 
\end{align} where $h$ is the pixel size. Such dependence on $h$ is not numerical stable for practical applications.

\subsection{Weighted Gaussian Curvature}
To tackle these problems in Gaussian curvature, we focus on weighted Gaussian curvature. More specifically, the weighted Gaussian curvature is defined as \cite{gong2013a}
\begin{equation}
\label{eq:wgc}
K^w(I(\vec{x}))=\frac{\mathrm{det}(D)}{\mathrm{det}(F)}=\frac{I_{xx}I_{yy}-I_{xy}^2}{1+I_x^2+I_y^2}\,.
\end{equation} We refer this equation as the classical scheme for computing weighted Gaussian curvature.

The weighted Gaussian curvature has several advantages over Gaussian curvature. First, weighted Gaussian curvature is more robust to image contrast change. Again, let $\hat{I}(\vec{x})=sI(\vec{x})$, where $s>0$. Then
\begin{equation}
K^w(\hat{I}(\vec{x}))=\frac{s^2(I_{xx}I_{yy}-I_{xy}^2)}{1+s^2(I_x^2+I_y^2)}\,.
\end{equation}At flat regions ($I_x^2+I_y^2\approx 0$), $K^w(\hat{I})\approx s^2K^w(I)$. But at sharp edges ($I_x^2+I_y^2>> 1$), $K^w(\hat{I})\approx K^w(I)$. Therefore, the weighted Gaussian curvature is robust with image contrast change at edges. In fact, the sharp edge regions are the main concern in various image processing tasks.

Second, as shown in following section, the weighted Gaussian curvature is independent from pixel size. Therefore, weighted Gaussian curvature can be used in the tasks where the pixel size is unknown.

Third, in this paper, a novel computing scheme for $K^w$ is developed. Our scheme does not require the image to be second order differentiable. Moreover, our scheme has high performance. The details are in the next section.

\section{A Discrete Scheme for Weighted Gaussian Curvature}
\label{sec:ds}
In this section, we will show a novel discrete scheme for weighted Gaussian curvature. Our scheme is based on the Gauss-Bonnet Theorem from differential geometry. Our scheme does not require the input image second order differentiable. Our scheme has a smaller support region and thus more accurate. Our scheme has a very high performance and can process 4K videos in real time.
\subsection{Gauss-Bonnet Theorem}
The well-known Gauss-Bonnet theorem in differential geometry states that
\begin{theorem}
	$\int_{\Psi}K\mathrm{d}\Psi+\int_{\partial\Psi}K_b\mathrm{d}b=2\pi \chi(\Psi)$\,,
\end{theorem} where $K_b$ is the boundary curvature, $\mathrm{d}b$ is boundary element and $\chi$ is the Euler characteristic.

Therefore, the integration of Gaussian curvature on the surface can be computed as
\begin{equation}
	\int_{\Psi}K\mathrm{d}\Psi=2\pi \chi(\Psi)-\int_{\partial\Psi}K_b\mathrm{d}b\,.
\end{equation}

\subsection{Discrete Gaussian Curvature}
\begin{figure}
	\centering
	\includegraphics[width=0.4\linewidth]{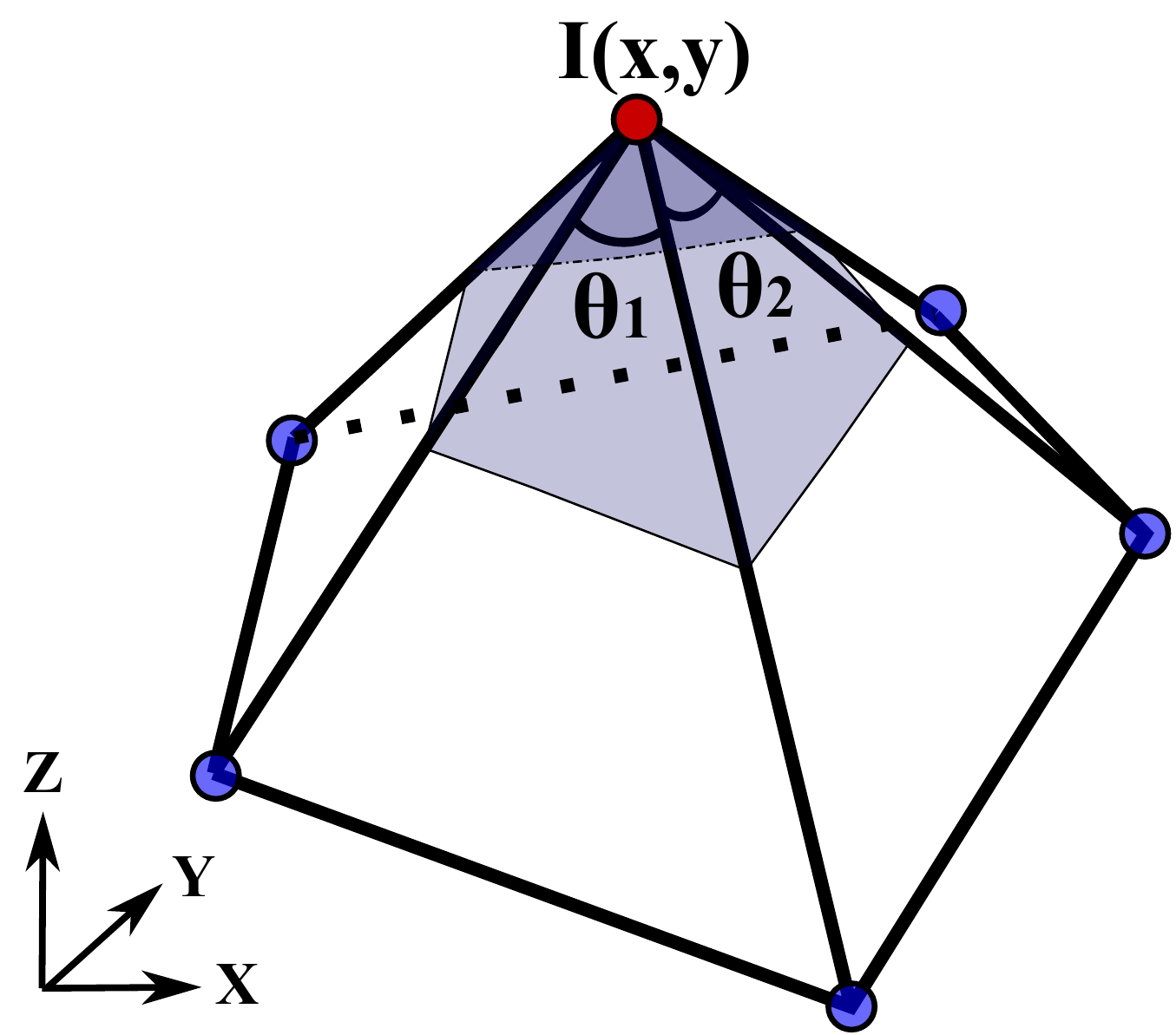}
	\caption{Discrete Gaussian Curvature on triangular meshes}
	\label{fig:mesh}
\end{figure}

When we restrict this equation on a local patch $A$, this equation becomes
\begin{equation}
\int_{A}K\mathrm{d}A=2\pi \chi(A)-\int_{\partial A}K_b\mathrm{d}b\,.
\end{equation} 
Above equation can lead to
\begin{equation}
K=\frac{2\pi \chi(A)-\int_{\partial A}K_b\mathrm{d}b}{\int_{A}\mathrm{d}A}\,,
\end{equation} which gives another way to estimate Gaussian curvature on the surface. On triangular meshes, this equation becomes \cite{article}
\begin{equation}
\label{eq:ds}
K=\frac{2\pi-\sum_{i}\theta_i(\vec{x})}{\mathrm{Area}(\vec{x})}\,,
\end{equation} where the $\theta_i$ is the angle around $\vec{x}$. Such geometric configuration is shown in Fig.~\ref{fig:mesh}

Although Eq.~\ref{eq:ds} does not require the surface to be second order differentiable, it is not clear how to compute the area related with $\vec{x}$ \cite{article}. To avoid this issue, we introduce the discrete weighted Gaussian curvature.
\subsection{Discrete Weighted Gaussian Curvature} 
Based on Eq.~\ref{eq:ds}, we can have the area weighted Gaussian curvature as following
\begin{equation}
	\label{eq:ours}
K^w=2\pi-\sum_{i}\theta_i(\vec{x})\,.
\end{equation}

This scheme is independent from spatial resolution, since the angle $\theta_i$ is not dependent on the triangle size but only depends on its shape. This property is important for scale adaptive image processing. We refer Eq.~\ref{eq:ours} as our method.
\subsection{Discrete Weighted Gaussian Curvature on Images}
\begin{figure}[!hbt]
	\centering
	\subfigure[four triangles for $I(x,y)$]{
		\label{fig:subfig:a} 
		\includegraphics[width=0.45\linewidth]{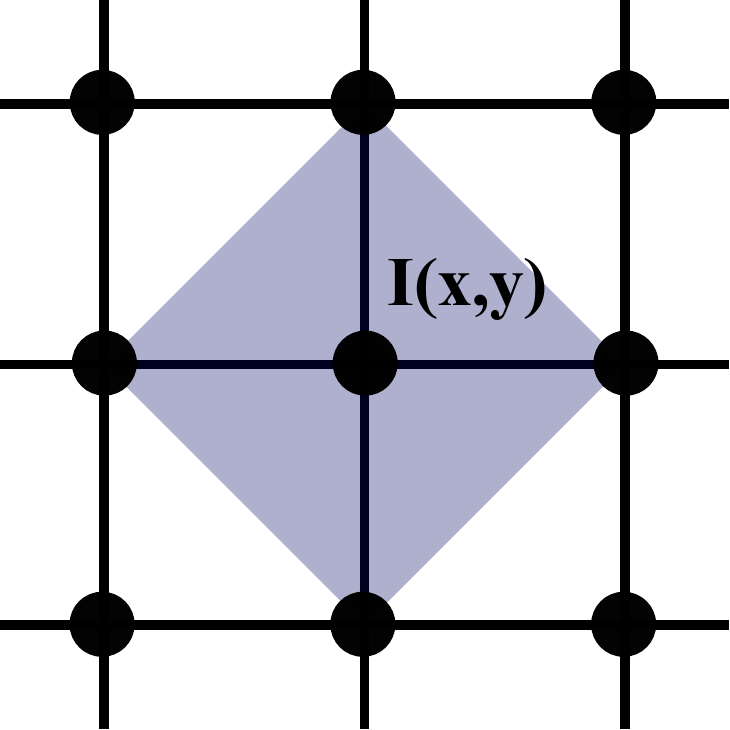}}
	\subfigure[the angle in one triangle]{
		\label{fig:subfig:b} 
		\includegraphics[width=0.45\linewidth]{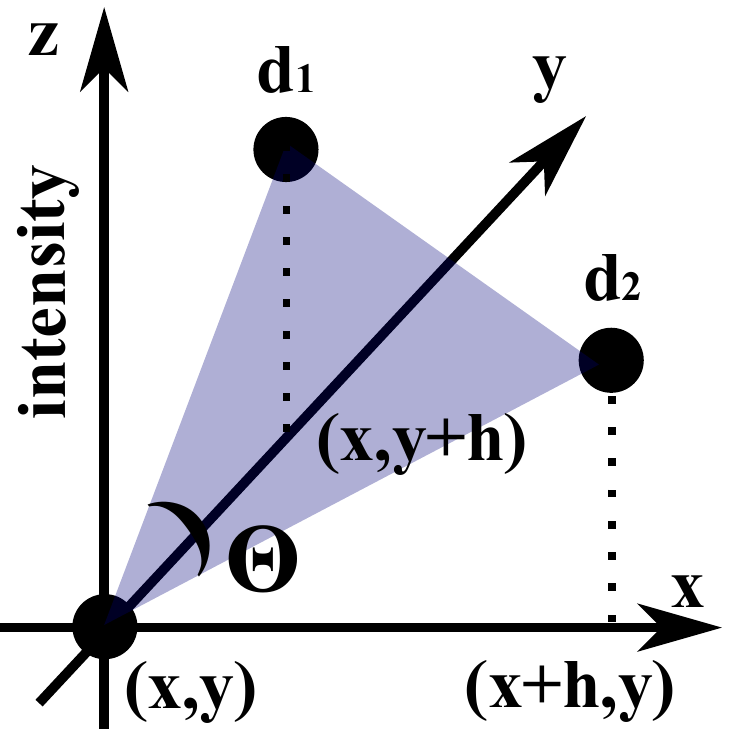}}
	\caption{Discrete Weighted Gaussian Curvature for Images}
	\label{fig:subfig} 
\end{figure}

Finally, we put Eq.~\ref{eq:ours} on discrete images that usually use Cartesian coordinates. More specifically, the center pixel $I(x,y)$ and its four neighbors $I(x-h,y)$, $I(x,y+h)$, $I(x+h,y)$ and $I(x,y-h)$ form four triangles. These four triangles are shown as shaded regions in Fig.~\ref{fig:subfig:a}.

In each triangle, we can put a local coordinate system at $(x,y,I(x,y))$. Then, the other vertexes of the triangle are located at $(h,0,I(x+h,y)-I(x,y))$ and $(0,h,I(x,y+h)-I(x,y))$, respectively. Such geometric configuration is shown in Fig.~\ref{fig:subfig:b}. For convenience, we define following scalars
\begin{align}
	d_{1}= &I(x,y+h)-I(x,y)\,, \\
	d_{2}= &I(x+h,y)-I(x,y)\,, \\
	d_{3}= &I(x,y-h)-I(x,y)\,, \\
	d_{4}= &I(x-h,y)-I(x,y)\,.
\end{align} Then, we can compute the angle $\theta_i$ by the cosine law
\begin{equation}
	\label{eq:angle}
	\theta_i=\arccos(\frac{d_id_{i+1}}{\sqrt{(h^2+d_i^2)(h^2+d_{i+1}^2)}})\,.
\end{equation}

\subsection{Acceleration by Lookup Table}
Thanks to the regular sampling on Cartesian grid and image intensity type (unsigned char), we can accelerate the computation of Eq.~\ref{eq:angle} by a lookup table.

\begin{figure}[!hbt]
	\centering
	\subfigure[Lookup Table for angle $\theta_i$]{	\includegraphics[width=0.48\linewidth]{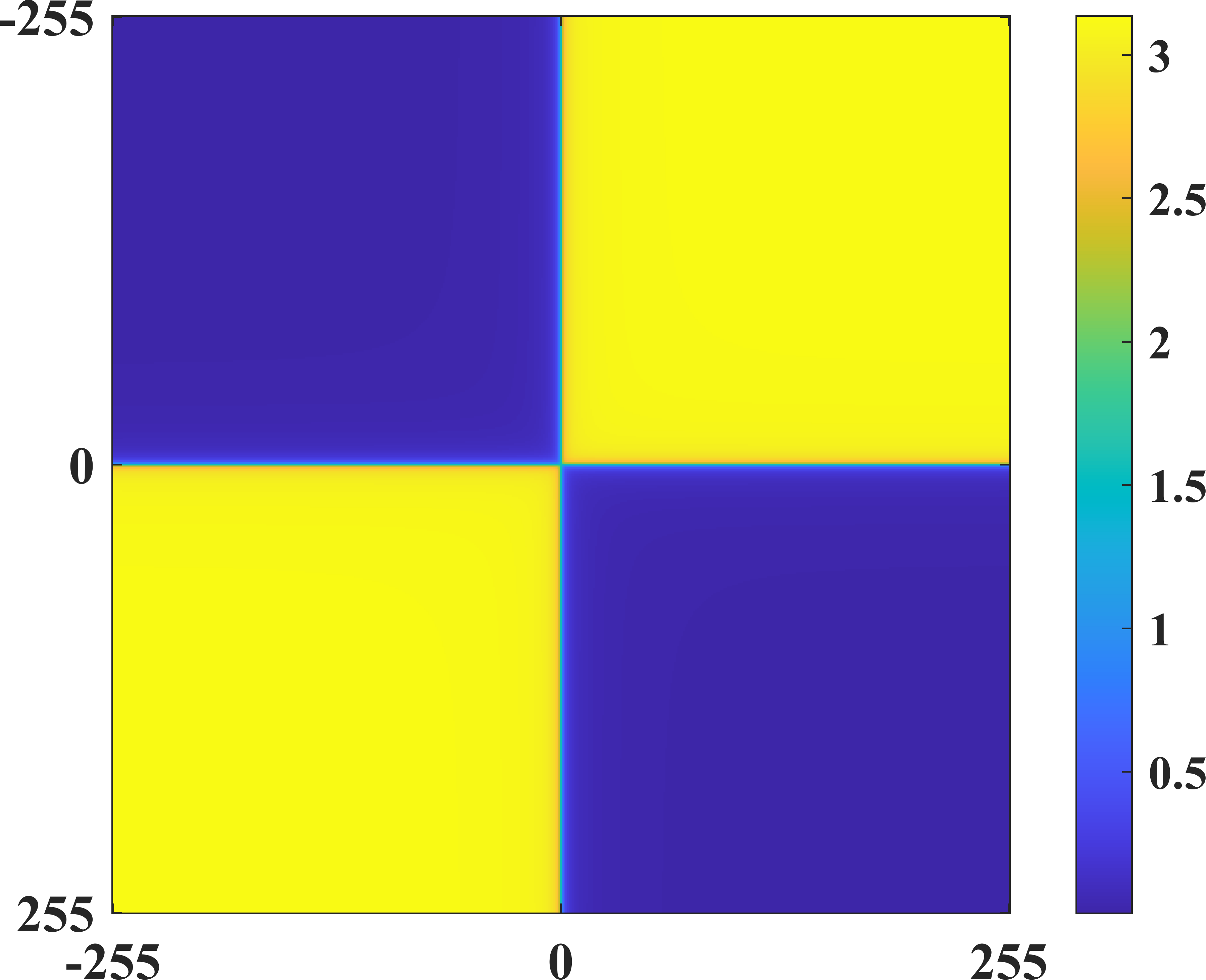}}
	\subfigure[the first row profile of the LUT]{	\includegraphics[width=0.48\linewidth]{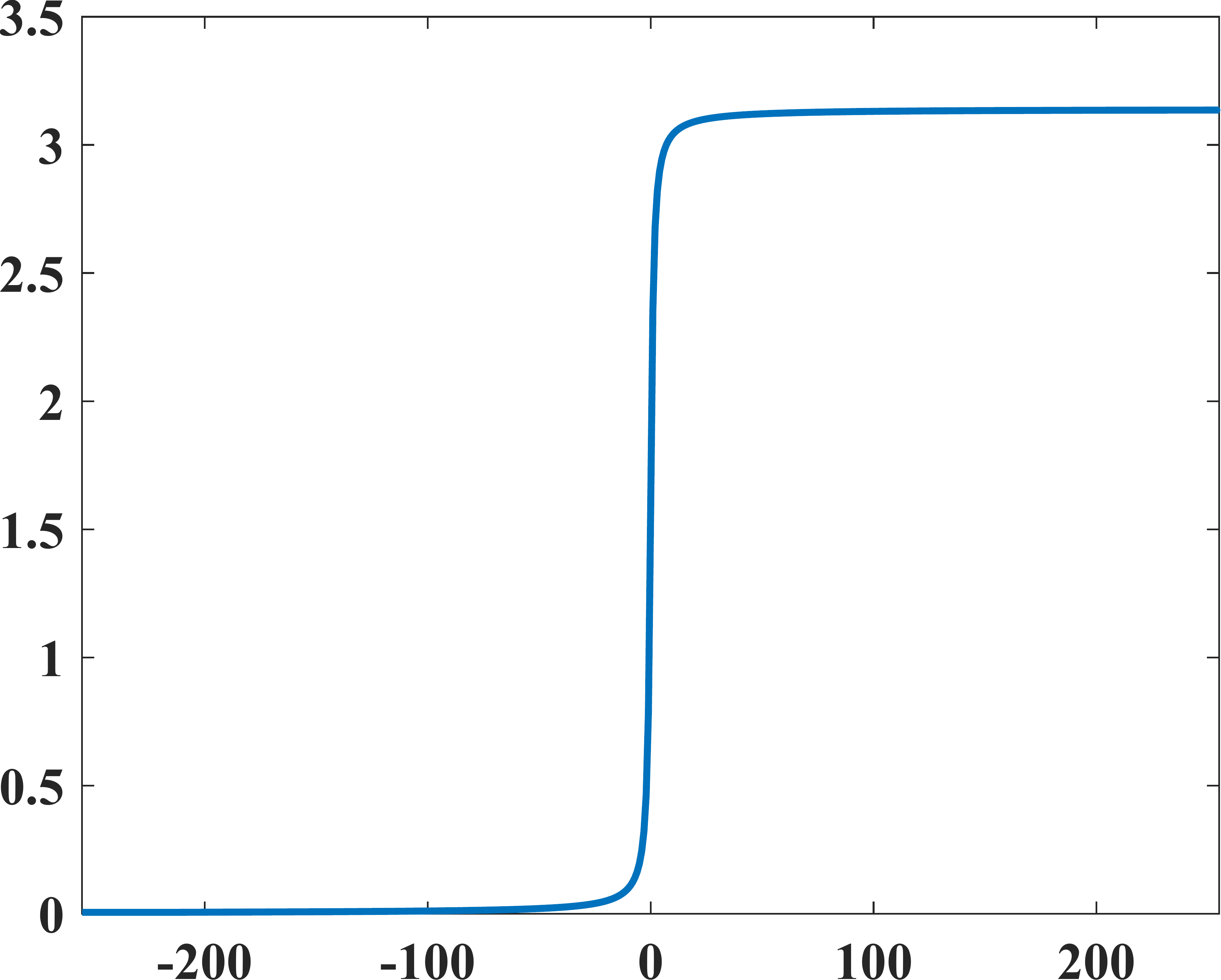}}
	\caption{The full lookup table (a) and its one line profile (b)}
	\label{fig:LUT}
\end{figure}

In most image processing tasks, we fix $h=1$. And the image intensity $I(x,y)$ is an integer in $[0,255]$. Therefore, $d_i$ are integers in $[-255,255]$. Since Eq.~\ref{eq:angle} only involves two independent variables, we can construct a lookup table to reduce the computation. The lookup table has size $511\times511$ and can be pre-computed for any images. The full lookup table is shown in Fig.~\ref{fig:LUT}(a).

The full lookup table is almost piece wise constant. Its first row ($d_i=0$) profile is shown in Fig.~\ref{fig:LUT}(b). Therefore, the size of such lookup table can be further reduced if storage or memory is restricted, for example, on embedded devices.

More specifically, when $|d_i|<T$ or $|d_{i+1}|<T$, the lookup table is used ($T$ is a threshold parameter). Other cases are approximated by constant values ($0$ or $\pi$), according to the sign of $d_id_{i+1}$.
\begin{equation}
	\theta_i=
	\begin{cases}
		\mathrm{LUT}(d_i,d_{i+1})&\mbox{if $|d_i|<T$ or $|d_{i+1}|<T$}\\
		\pi&\mbox{otherwise, $d_id_{i+1}>0$}\\
		0&\mbox{otherwise, $d_id_{i+1}<0$}
	\end{cases}\,.
\end{equation} In practice, the threshold value $T$ can be set to $31$. The reason comes from the gradient statistics of natural images \cite{gong:gdp,gong:phd}. Such partial lookup table only contains $2\times511\times(2T-1)-(2T-1)^2$ elements, which is much less than $511\times511$.
\subsection{Accuracy and Performance}
\begin{figure}
	\centering
	\subfigure[synthetic image $I$]{\includegraphics[width=0.35\linewidth]{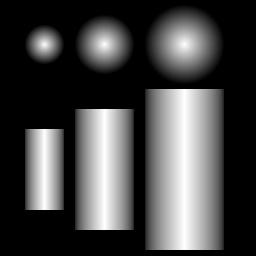}}
	\subfigure[its corresponding $\Psi=(\vec{x},I(\vec{x}))$]{\includegraphics[width=0.56\linewidth]{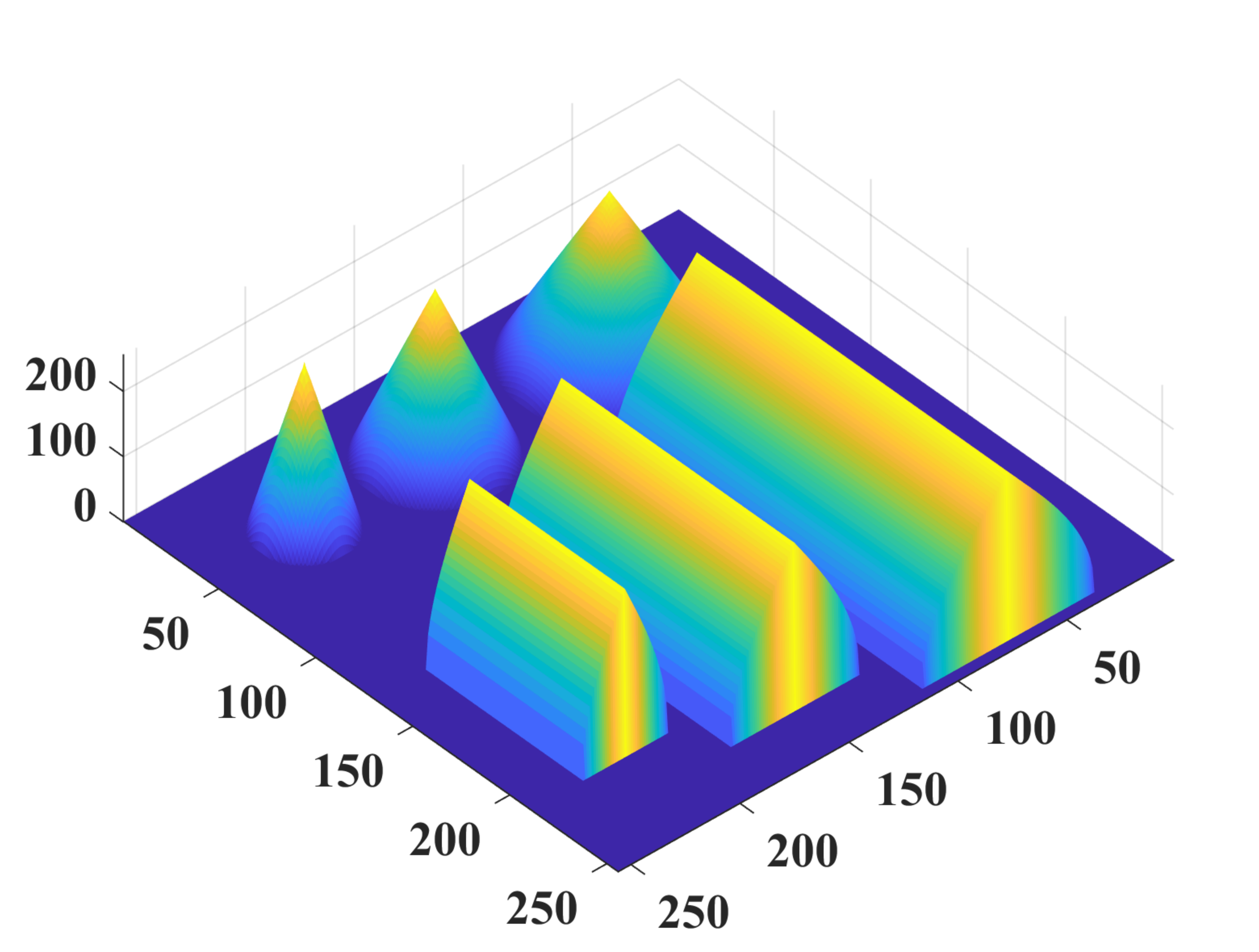}}
	\caption{One synthetic image and its surface representation}
	\label{fig:surf}
\end{figure}

To compare our method with the classical computation scheme (Eq.~\ref{eq:wgc}), we perform both of them on a synthetic image, which only contains cones and cylinder structure, as shown in Fig.~\ref{fig:surf}(a). Theoretically, the weighted Gaussian curvature on such image is zero. Therefore, the mean of absolute weighted Gaussian curvature indicates the numerical error.

\begin{figure}
	\centering
	\subfigure[$K^w$ by Eq.~\ref{eq:wgc}]{\includegraphics[width=0.4\linewidth]{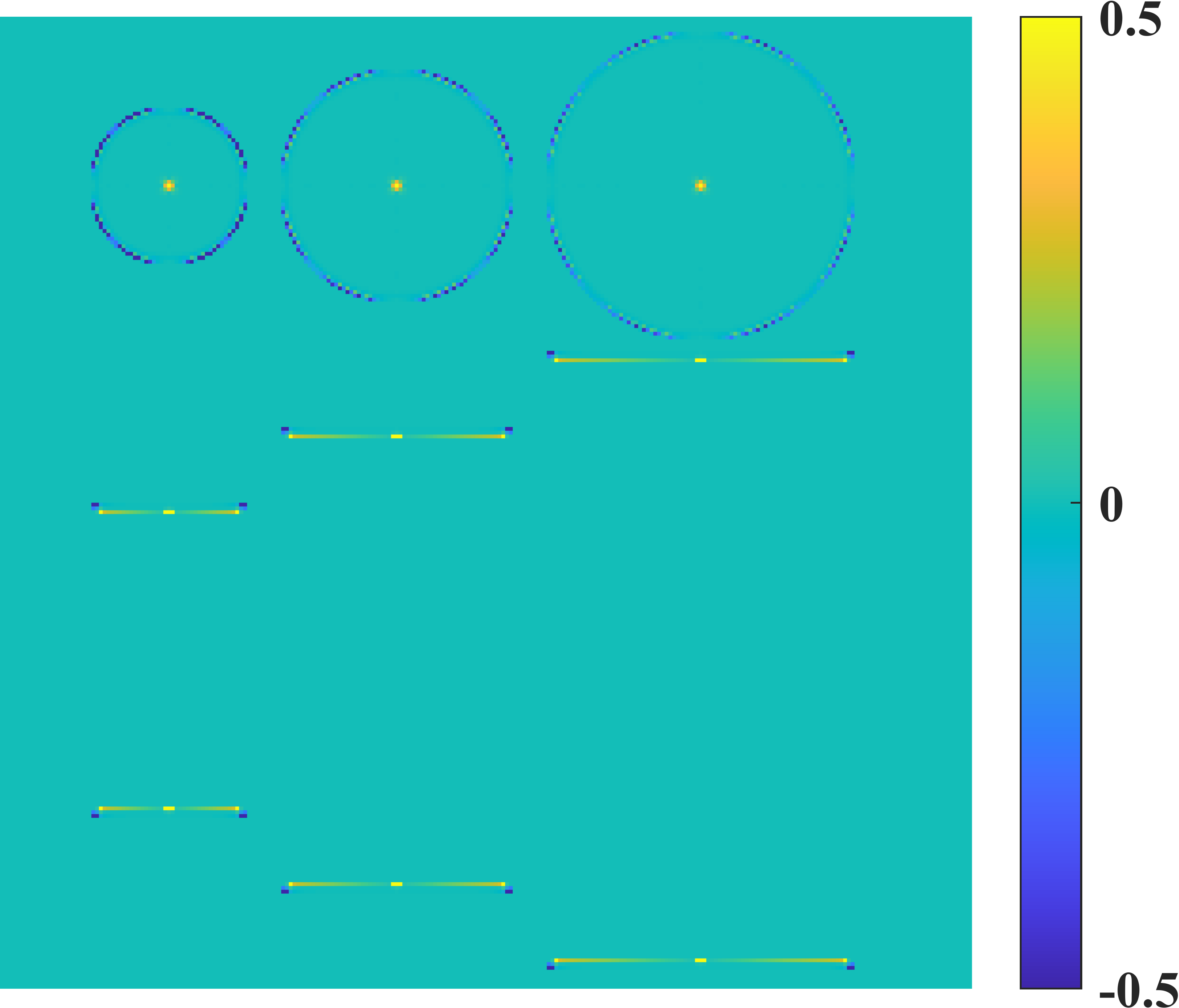}}
	\subfigure[$K^w$ by our method]{\includegraphics[width=0.4\linewidth]{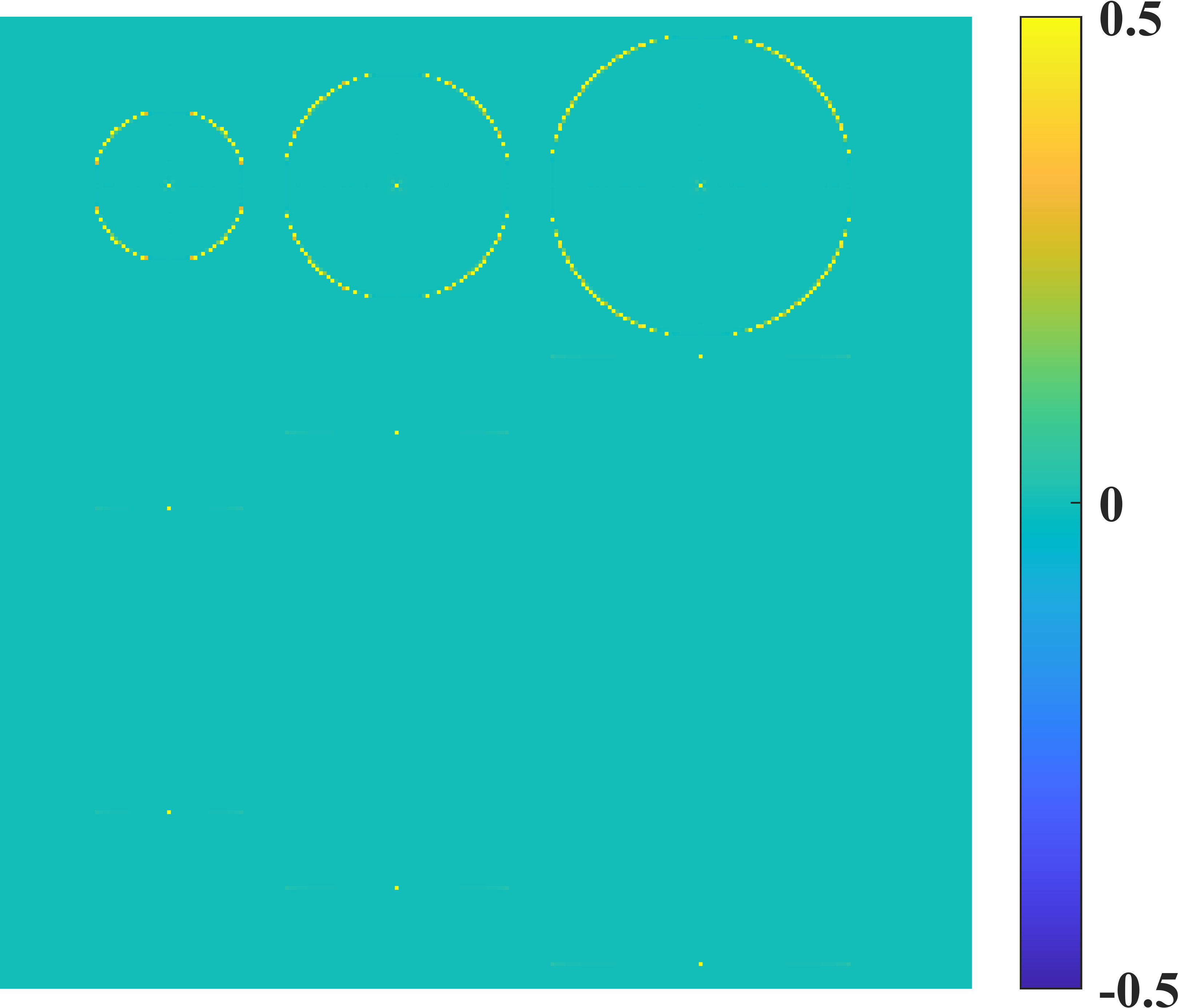}}
	\caption{The classical scheme (a) and our method on the synthetic image. Ideally, the mean of $|K^w|$ should be zero on this image. The mean of $|K^w|$ in (a) and (b) is 0.065 and 0.0038, respectively. This indicates that our method is more accurate.}
	\label{fig:syn}
\end{figure}

We perform the classical scheme (Eq.~\ref{eq:wgc}) and our method on this synthetic image. The results are shown in Fig.~\ref{fig:syn}. The mean of $|K^w|$ in Fig.~\ref{fig:syn}(a) and (b) is 0.065 and 0.0038, respectively. This indicates that our method is more accurate.

\begin{table}[hbt]
	\centering
	\caption{Comparison between Eq.~\ref{eq:wgc} and Eq.~\ref{eq:ours}} 
	\label{table1}  
	\begin{tabular}{|c|c|c|}  
		\hline  
		& classical scheme & our method\\
		\hline
		support region & full window& cross  \\
		\hline
		differential & second order& zero order \\
		\hline
		accuracy & low& high \\
		\hline
		performance & low& very high \\
		\hline
	\end{tabular}
\end{table}

The reason is that our method has a smaller support region. From geometric point of view, the smaller support region, the better estimation of weighted Gaussian curvature. It is clear that our method has a $3\times3$ cross support region. In contrast, the classical scheme (Eq.~\ref{eq:wgc}) has a $3\times3$ full window support region. Therefore, our method can capture more local geometric information. A comparison between our method and the classical scheme is given in Table~\ref{table1}.

Our method is implemented with CUDA C++ language and performed on a GTX 1080 Ti GPU card (3584 CUDA cores, 11GB memory). The full lookup table is put in the constant memory. Our method can achieve 890 frames per second for 4K videos. 
\subsection{More Results on Real Images}
\begin{figure}
	\centering
	\subfigure[input image]{\includegraphics[width=0.32\linewidth]{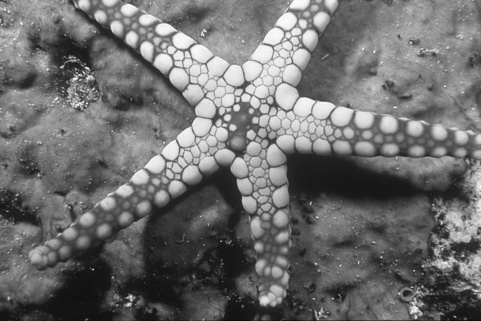}}
	\subfigure[$K^w$ by Eq.~\ref{eq:wgc}]{\includegraphics[width=0.32\linewidth]{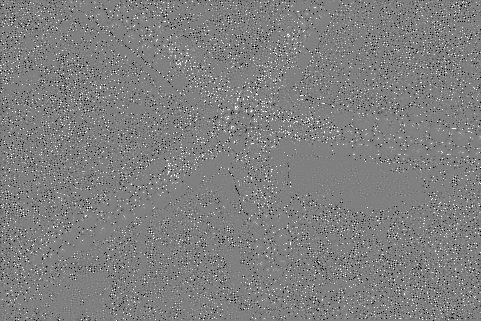}}
	\subfigure[$K^w$ by our method]{\includegraphics[width=0.32\linewidth]{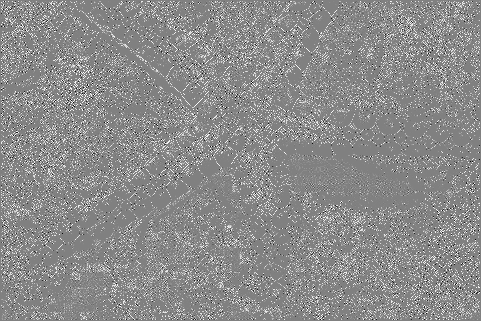}}
	
	\subfigure[input image]{\includegraphics[width=0.32\linewidth]{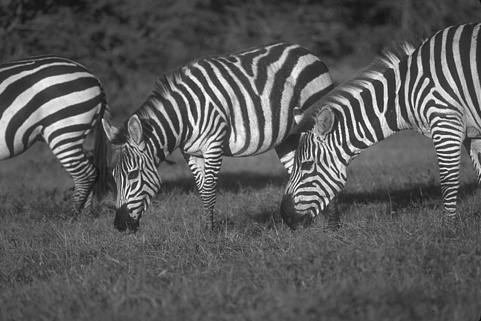}}
	\subfigure[$K^w$ by Eq.~\ref{eq:wgc}]{\includegraphics[width=0.32\linewidth]{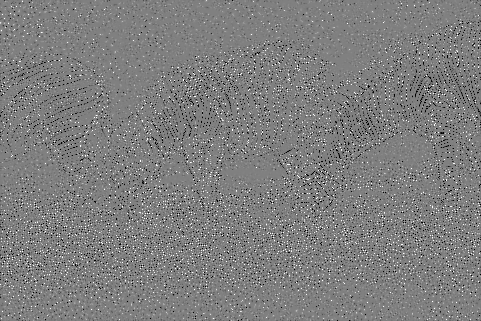}}
	\subfigure[$K^w$ by our method]{\includegraphics[width=0.32\linewidth]{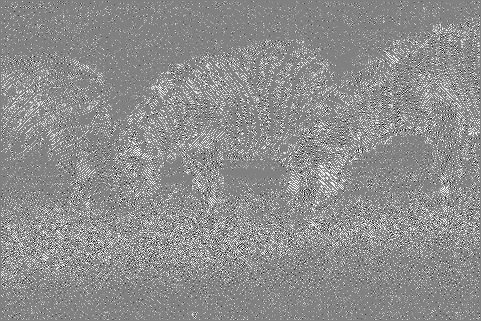}}
	
		\subfigure[input image]{\includegraphics[width=0.32\linewidth]{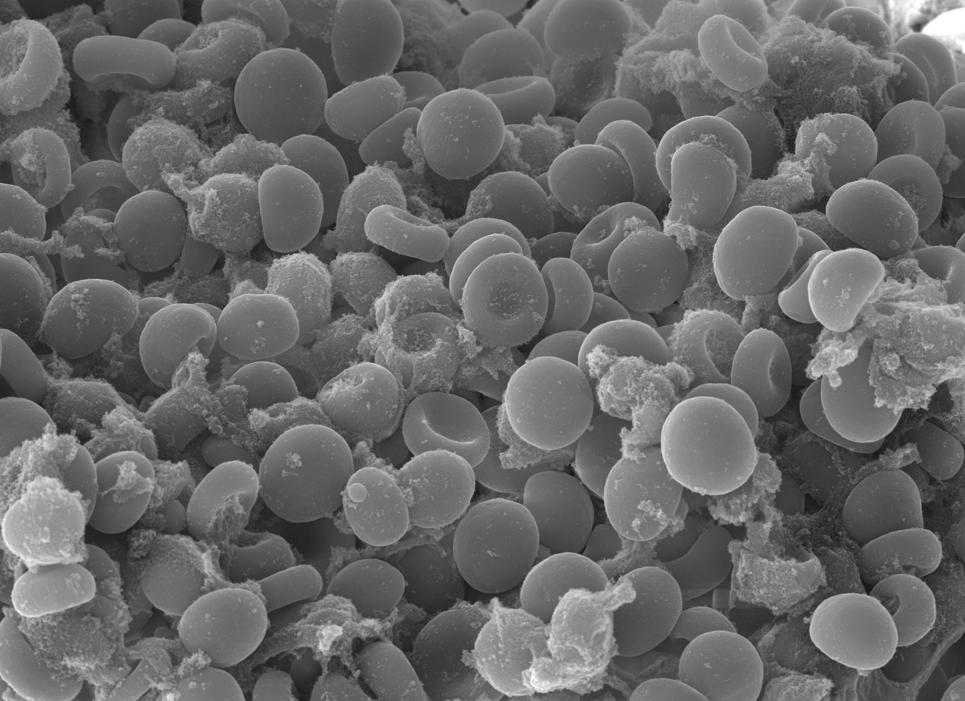}}
	\subfigure[$K^w$ by Eq.~\ref{eq:wgc}]{\includegraphics[width=0.32\linewidth]{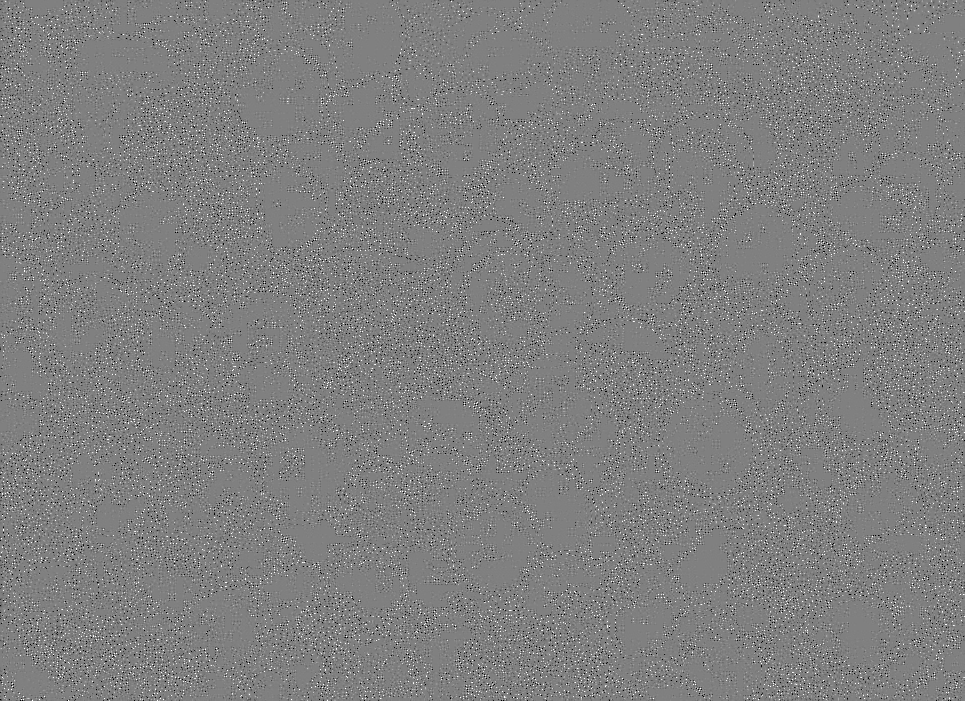}}
	\subfigure[$K^w$ by our method]{\includegraphics[width=0.32\linewidth]{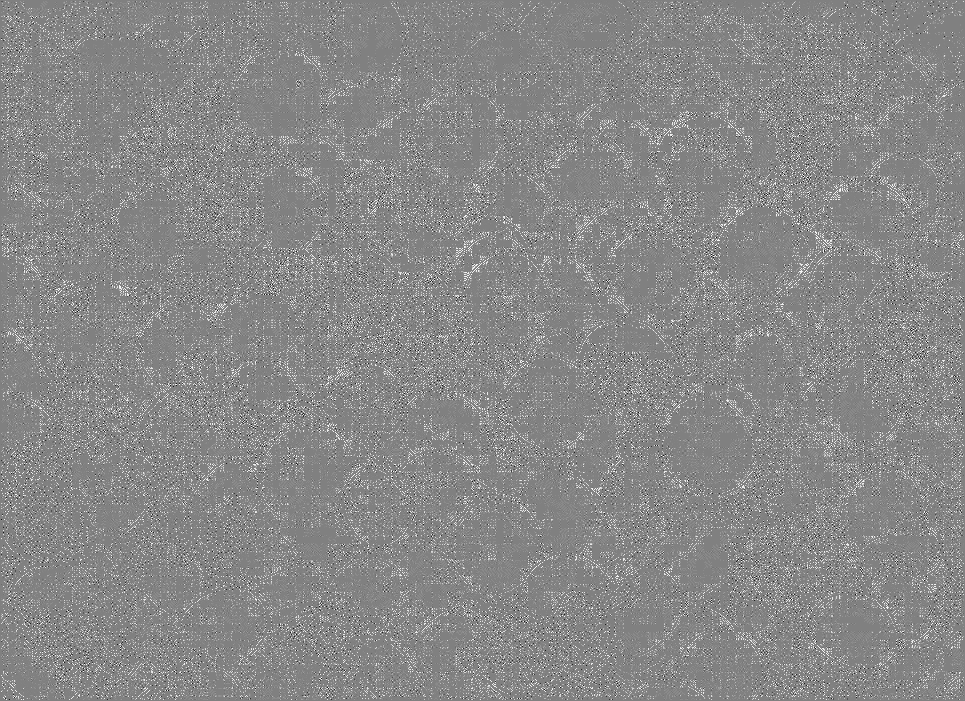}}
	\caption{The input image (left), the $K^w$ by classical scheme (middle) and $K^w$ by our method (right). Both are shown by $128+20K^w$ for better visualization. Our method can capture more structural information.}
	\label{fig:exp}
\end{figure}

Our method and the classical scheme are tested on two natural images and one biomedical image, as shown in Fig.~\ref{fig:exp}. It can be told that our method captures more local and geometrical information, thanks to its smaller support region. The repeated pattern in the input images can be seen from our results. Moreover, thanks to the lookup table technique, our method can compute weighted Gaussian curvature for seven billion pixels in one second, which is much faster than the classical scheme.

\section{Conclusion}
In this paper, we propose a novel computation scheme for weighted Gaussian curvature on images. Our method does not require the input image to be second order differentiable. Our method has a smaller support region than the classical scheme. Therefore, it is more accurate in capturing local geometric information. Moreover, thanks to the lookup table technique, its numerical performance is very high. Our scheme can be adopted in a large range of applications, where the weighted Gaussian curvature is needed, such as image reconstruction \cite{hao2017scattered,Gong2012}, motion estimation \cite{Dai:2016}, image enhancement \cite{Gong:2014a,Fei:2015}, medical image analysis \cite{Gong2019,Liu2019,Gong2018,Yin2019}, etc.

\bibliographystyle{IEEEbib}
\bibliography{IP}

\end{document}